\documentclass{article} 

\PassOptionsToPackage{numbers, compress}{natbib}
\usepackage[final]{neurips_2022}

\usepackage{microtype}
\usepackage{graphicx}
\usepackage{subfigure}
\usepackage{booktabs} 

\usepackage{url}
\usepackage[utf8]{inputenc} 
\usepackage[T1]{fontenc}    
\usepackage{url}            
\usepackage{booktabs}       
\usepackage{amsfonts}       
\usepackage{nicefrac}       
\usepackage{microtype}      
\usepackage{amsmath}
\usepackage{graphicx}
\usepackage{epigraph}
\usepackage[export]{adjustbox}
\usepackage[final]{pdfpages}
\usepackage{mathrsfs}
\usepackage{makecell}
\usepackage{enumitem}
\usepackage{pdfpages}
\usepackage{amssymb}
\usepackage{longtable}
\usepackage{multirow}
\usepackage{wrapfig}
\usepackage{tcolorbox}
\usepackage{tabu}
\usepackage{authblk}
\usepackage{algorithm}
\usepackage{xcolor}
\usepackage{algpseudocode}
\usepackage[title]{appendix}

\usepackage{listings}
\definecolor{codegreen}{rgb}{0,0.6,0}
\definecolor{codegray}{rgb}{0.5,0.5,0.5}
\definecolor{codepurple}{rgb}{0.58,0,0.82}
\definecolor{backcolour}{rgb}{0.95,0.95,0.92}
\lstdefinestyle{pythonstyle}{
    backgroundcolor=\color{backcolour},   
    commentstyle=\color{codegreen},
    keywordstyle=\color{magenta},
    numberstyle=\tiny\color{codegray},
    stringstyle=\color{codepurple},
    basicstyle=\ttfamily\footnotesize,
    breakatwhitespace=false,         
    breaklines=true,                 
    captionpos=b,                    
    keepspaces=true,                 
    numbers=left,                    
    numbersep=5pt,                  
    showspaces=false,                
    showstringspaces=false,
    showtabs=false,                  
    tabsize=2
}
\newcommand{\customfootnotetext}[2]{{
  \renewcommand{\thefootnote}{#1}
  \footnotetext[0]{#2}}}

\usepackage{amsmath,amsfonts,bm}

\def\eqref#1{equation~\ref{#1}}

\def\1{\bm{1}}

\def\vb{{\bm{b}}}
\def\vc{{\bm{c}}}

\def\vf{{\bm{f}}}
\def\vg{{\bm{g}}}
\def\vh{{\bm{h}}}
\def\vi{{\bm{i}}}

\def\vz{{\bm{z}}}

\def\mK{{\bm{K}}}

\def\mQ{{\bm{Q}}}

\def\mV{{\bm{V}}}
\def\mW{{\bm{W}}}

\DeclareMathAlphabet{\mathsfit}{\encodingdefault}{\sfdefault}{m}{sl}
\SetMathAlphabet{\mathsfit}{bold}{\encodingdefault}{\sfdefault}{bx}{n}

\newcommand{\sigmoid}{\sigma}

\usepackage{hyperref}       

\title{\center{Block-Recurrent Transformers}}

\author{\textbf{DeLesley Hutchins$^{*1}$, Imanol Schlag$^{*3\dagger}$, Yuhuai Wu$^1$,  Ethan Dyer$^2$, Behnam Neyshabur$^2$}\\
$^1$ Google Research\qquad
$^2$ Google Research, Blueshift Team \\
$^3$ The Swiss AI Lab IDSIA, SUPSI \& USI \\
\texttt{\{delesley, yuhuai, edyer, neyshabur\}@google.com\qquad
\texttt{imanol@idsia.ch}}
}

\newcommand{\namett}[1]{{\texttt{\small #1}}}

\begin{document}

\maketitle

\customfootnotetext{$*$}
{Equal Contribution\qquad
{$\dagger$}
Work done partially while interning at Google Research (Blueshift Team) and partially funded by ERC Advanced grant no: 742870 to J.Schmidhuber.}

\begin{abstract}
We introduce the Block-Recurrent Transformer, which applies a transformer layer in a recurrent fashion along a sequence, and has linear complexity with respect to sequence length.  Our recurrent cell operates on blocks of tokens rather than single tokens during training, and leverages parallel computation within a block in order to make efficient use of accelerator hardware.  The cell itself is strikingly simple. It is merely a transformer layer: it uses self-attention and cross-attention to efficiently compute a recurrent function over a large set of state vectors and tokens.  Our design was inspired in part by LSTM cells, and it uses LSTM-style gates, but it scales the typical LSTM cell up by several orders of magnitude.  Our implementation of recurrence has the same cost in both computation time and parameter count as a conventional transformer layer, but offers dramatically improved perplexity in language modeling tasks over very long sequences. Our model out-performs a long-range Transformer XL baseline by a wide margin, while running twice as fast. We demonstrate its effectiveness on PG19 (books), arXiv papers, and GitHub source code. Our code has been released as open source \cite{meliad2022github}.
\end{abstract}

\section{Introduction}

Transformers have mostly replaced recurrent neural networks (RNNs), such as LSTMs \cite{hochreiter97lstm}, on tasks that involve sequential data, especially natural language.
There are several reasons for their success. 
First, transformers process all elements of the sequence in parallel, and are thus faster to train on modern accelerator hardware.  In contrast, an RNN must process tokens sequentially, which leads to slow step times during training, and large batch sizes in order to fully saturate GPUs or TPUs.

Second, an RNN must summarize and compress the entire previous sequence into a single state vector which is passed from one token to the next.  The size of the state vector limits the amount of information that the RNN can encode about the previous tokens in the sequence.  In contrast, a transformer can attend directly to past tokens, and does not suffer from this limitation.

Third, attention operates effectively over longer distances.  The forget gate in an LSTM discards information moving forward, and causes vanishing gradients during backpropagation. In practice, this means that LSTMs struggle to send a clear signal over more than a few hundred tokens, far less than the typical size of the attention window in a transformer~\cite{khandelwal-etal-2018-sharp}.  

Despite these advantages, transformers also have a disadvantage.  The computational complexity of self-attention is quadratic with respect to the sequence length, which is a limiting factor when attempting to process long documents, such as books, technical articles, or source code repositories. Moreover, a transformer has no memory of past context; any tokens that it cannot attend to are ``invisible'' to the model.

In this work, we describe an architecture which combines the benefits of attention and recurrence.  Like previous implementations of recurrence, our architecture constructs and maintains a fixed-size state, which summarizes the sequence that the model has seen thus far.  However, our implementation of recurrence differs from previous work in several important aspects which together address the three limitations mentioned above.

\begin{figure}[t]
    \vskip -3ex
    \includegraphics[width=5.5in]{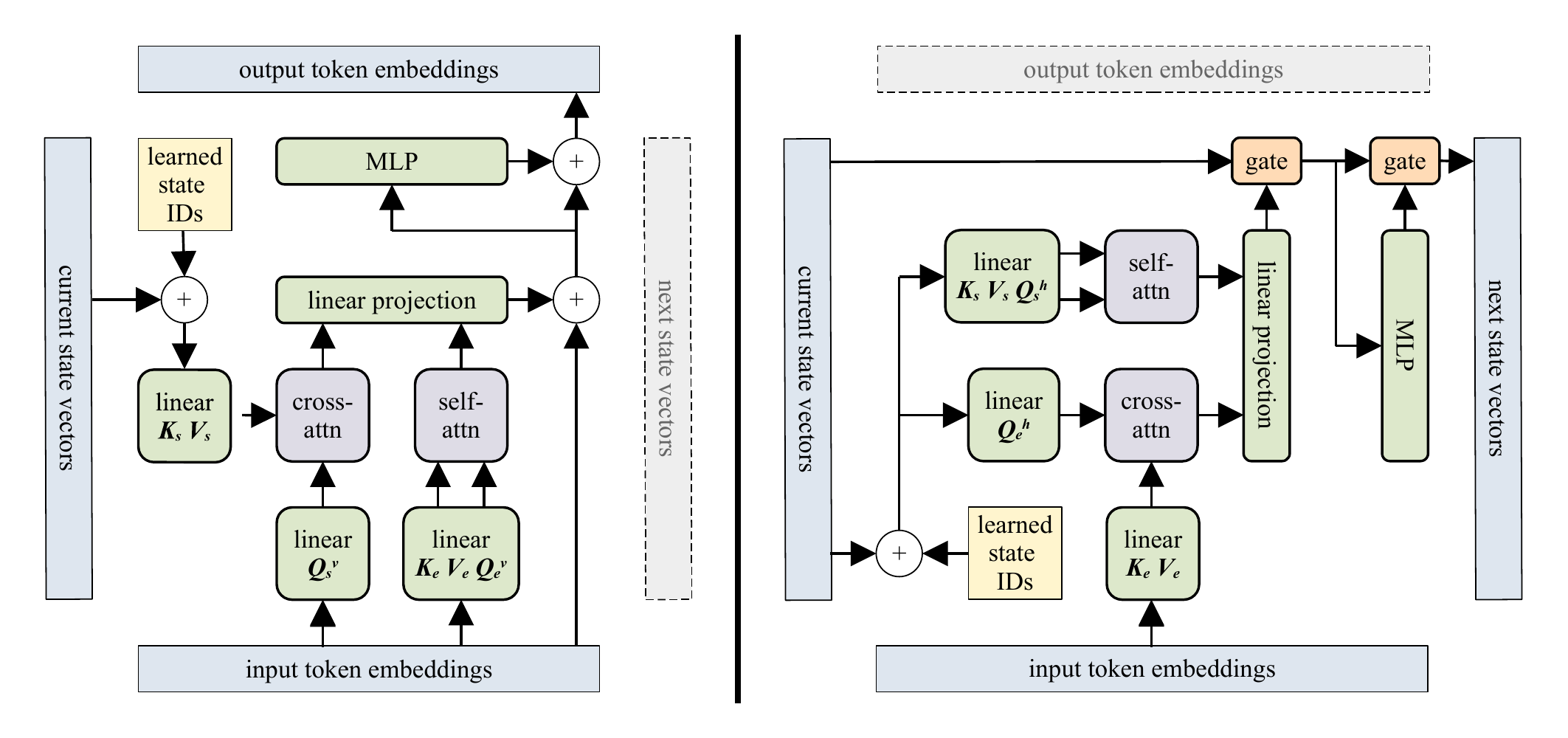}
    \vskip -2ex
    \caption{Illustration of our recurrent cell. The left side depicts the vertical direction (layers stacked in the usual way) and the right side depicts the horizontal direction (recurrence).  Notice that the horizontal direction merely rotates a conventional transformer layer by 90$^\circ$, and replaces the residual connections with gates.}
    \label{fig:recurrent-cell}
    \vskip -2ex
\end{figure}

Instead of processing the sequence one token at a time, \textbf{our recurrent cell operates on \emph{blocks} of tokens}; see Figure \ref{fig:recurrent-cell}. Within a block, all tokens are processed in parallel, at least during training.  The recurrent cell likewise \textbf{operates on a \emph{block} of state vectors rather than a single vector.} This means that the size of the recurrent state is orders of magnitude larger than in an LSTM, which dramatically improves the model's capacity to capture the past.  Processing the sequence in blocks also helps propagate information and gradients over longer distances, because the number of recurrent steps (and thus the number of times that the forget gate is applied) is orders of magnitude smaller.  We show that the Block-Recurrent Transformer can remember information over distances of 60k tokens or more.
\looseness=-1

The recurrent cell itself is strikingly simple.  For the most part, it consists of an ordinary transformer layer applied in a recurrent fashion along the sequence length.  There are \textbf{a few tricks that are necessary to stabilize training}; see Sections \ref{section:state-ids} and \ref{section:gate-initialization} for details.
The cost of recurrence, in terms of both computation time and parameter count, is essentially the same as simply adding one more layer to our transformer baseline. We demonstrate empirically that adding a single recurrent layer results in a much larger improvement in perplexity on multiple datasets than adding a conventional transformer layer, while training time and memory use are equivalent.
Moreover, our recurrent cell is very easy to implement because it largely makes use of existing transformer code.  Thus, our technique is a cheap and cheerful way to improve language modeling perplexity on long sequences.

\section{Related Work}
\label{section:related-work}

The quadratic cost of attention is well known in the literature, and a great deal of work has been done on efficient long-range attention mechanisms; see \cite{tay2021long, tay2020efficient} for recent surveys. 
Sparse strategies such as Big Bird \cite{Zaheer2020bigbird}, Routing Transformers \cite{roy2021efficient}, and Reformer \cite{kitaev2020reformer} select only a subset of tokens to attend to. Hierarchical mechanisms \cite{Ainslie2020etc_hierarchical_structured} combine multiple tokens into phrases or sentences to reduce sequence length.
Expire-span \cite{sukhbaatar2021expire} learns to prune far-away tokens that the model has labelled as ``unimportant''.  
Memorizing transformers \cite{wu2022memorizing} replace dense attention with  $k$-nearest-neighbor lookup.

Yet another approach is to reduce the sequence length by pooling, averaging, or compressing it in some way. Hierarchical 1D attention \cite{Zhu2021hierarchical1D}, and Combiner \cite{ren2021combiner} apply pooling or averaging over tokens at longer distances.  Linformer \cite{wang2020linformer} applies a linear transformation to the key and value matrices to reduce the sequence length. Compressive transformers \cite{rae2020compressive} and funnel transformers \cite{dai2020funnel} apply additional learned compression layers to compress the sequence.

The equation for attention is (roughly) $\textsf{softmax}(\mQ\mK^T) \mV$ where $\mQ, \mK,$ and $\mV$ are the query, key, and value matrices of the attention layer.  If the softmax operation is removed from this equation or somehow ``linearized'', the equation can be rearranged as $\mQ (\mK^T\mV)$, where $(\mK^T\mV)$ can be computed incrementally (i.e., in a recurrent fashion) as a cumulative sum over the sequence \cite{katharopoulos2020transformersRNNs}.  Linearized attention thus has linear rather than quadratic complexity with respect to sequence length.
Following this line of reasoning, there have been several proposals that approximate the softmax \cite{choromanski2021performers, peng2021random} or replace it \cite{schlag2021linear, Hua2022lineartime}.  Linear transformers are related to earlier work on fast weight programmers \cite{schlag2021linear} \cite{schmidhuber1993reducing}, and can be extended with other forms of recurrence \cite{irie2021going}.


Our work differs from all of the above mechanisms, because we rely only on standard dense attention with softmax.

A few other lines of research have combined the transformer architecture with recurrence in some way.
The feedback transformer \cite{fan2020feedback} allows lower layers to attend to the output of the topmost layer.  
Feedback has minimal cost at inference time, but it is unfortunately very slow to train because tokens must be processed sequentially.  
Simple Recurrent Units \cite{lei18sru, lei21srupp} use a recurrence function that does not involve matrix multiplication, and is consequently much faster.
RNMT+ combines RNNs and transformers in an encoder/decoder architecture to improve on translation tasks \cite{chen2018bestRNMT}.  ``Sandwich models'' alternate between transformer and RNN layers and out-perform both transformers and RNNs on tasks involving source code \cite{hellendoorn2020great}.  The R-Transformer introduces an additional local RNN which can be computed in parallel in order to better model sequential structure \cite{wang2020rtransformer}.  The Perceiver architecture \cite{jaegle21perceiver} is somewhat similar to ours; it also applies a transformer layer in an iterative fashion.

To the best of our knowledge, the idea of performing recurrence on blocks of tokens is underexplored. In the context of translation, \cite{adel2021memorytransformer} operates on sentences rather than tokens.
Staircase Attention \cite{ju2021staircase} also operates on blocks of tokens; each layer takes, as input, the outputs of the same layer from the previous block.  

\section{Method}

The Block-Recurrent Transformer is based on sliding-window attention \cite{beltagy2020longformer}, which is an extension of ideas from Transformer-XL \cite{dai2019transformerXL}. 
  
A long document, such as a book, consists of a \emph{sequence} of tokens.  Due to memory limitations, it is usually not possible to fit the entire sequence into device memory.  Thus, the sequence is divided into \emph{segments} of length $N$ ($N = 4096$ in our experiments), which are processed sequentially over a number of training steps.  Each training step processes one segment.  

\begin{figure}[t]
\begin{center}
\vskip -3ex
\includegraphics[width=2.75in]{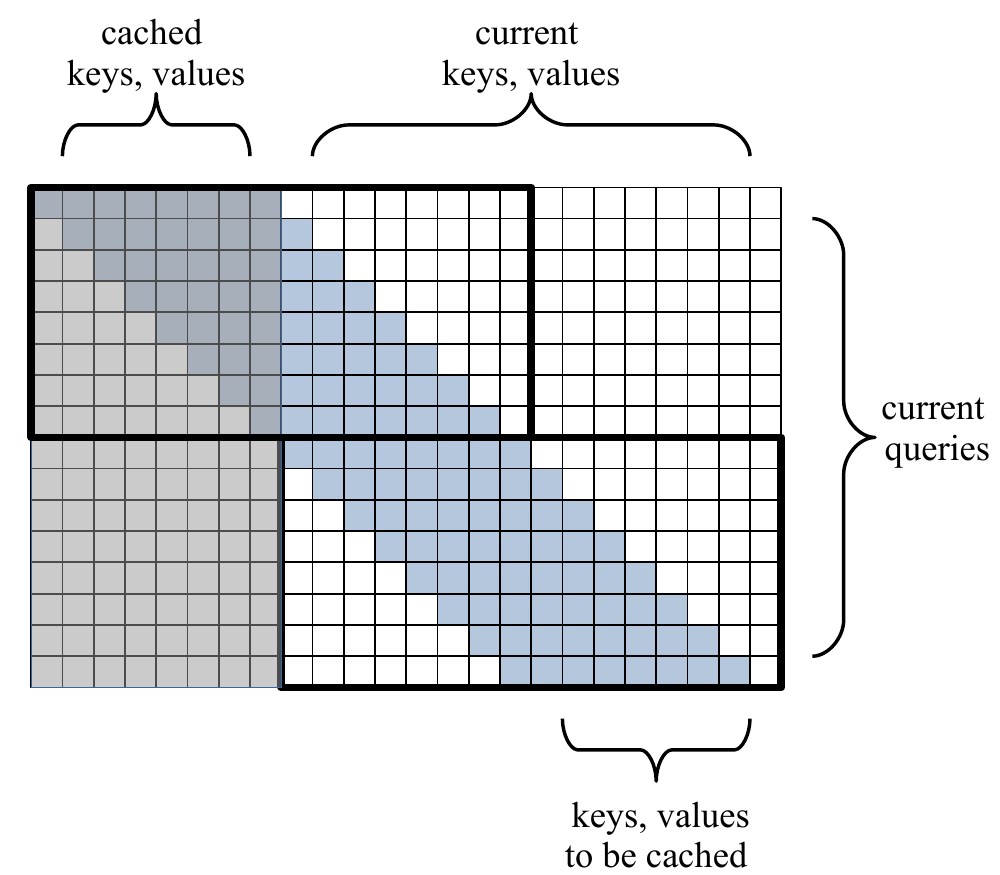}
\caption{Sliding window, where segment length $N = 16$, window/block size $W = 8$. Keys and values for the first $W$ shaded tokens were computed and cached on the previous training step; the remaining $N$ unshaded tokens are the segment for the current training step.  Instead of a single $N \times (W + N)$ attention matrix, attention is done in two tiles of size $W \times 2W$.}
\label{fig:sliding-window}
\end{center}
\vskip -3ex
\end{figure}

The sliding window attention pattern is illustrated in Figure \ref{fig:sliding-window}. Given a segment of $N$ tokens, the sliding window applies a causal mask in which each token can only attend to the $W$ previous tokens, where $W$ is the \emph{window size} ($W = 512$ in our experiments).  Because of the causal mask, most entries of the $N \times N$ attention matrix are masked out (assuming that $W << N$).  Thus, the attention computation can be optimized by breaking it into smaller tiles along the diagonal.  The segment of $N$ tokens is subdivided into \emph{blocks} of size $W$, and each block attends locally to itself and to the previous block, so the size of each local attention matrix is $W \times 2W$. Using this mechanism, attention is quadratic with respect to the window size $W$, but linear with respect to the segment length $N$. 

Borrowing an idea from Transformer-XL, the keys and values from the last block in each segment are stored in a non-differentiable \emph{cache} for use on the next training step.  By using the cache, the first block in the next segment can attend to the last block in the previous segment, which extends the sliding window to cover the entire (book-length) sequence.  The cache implements a form of truncated backpropagation through time \cite{Williams1990tbptt} over long documents.  

Note that if $N = W$, then sliding window attention will behave exactly like Transformer-XL; it will process and cache one segment (i.e. one block) per training step.  Setting $N >> W$ does not change the context length of attention, but it allows gradients to backpropagate across multiple blocks during training; we show that the improved differentiability provides a modest benefit to perplexity over Transformer-XL.  See Appendix \ref{appendix:sliding-attention} for more details.

\subsection{Recurrent Cell}
 
A Block-Recurrent Transformer layer extends the sliding-window attention mechanism by adding a set of recurrent states, which are updated at the end of each block of $W$ tokens. Our design for the recurrent cell is illustrated in Figure \ref{fig:recurrent-cell}, which depicts the operations done within a single block of the input sequence.

The recurrent cell receives two tensors as inputs: a set of $W$ token embeddings, where $W$ is the block/window size, and a set of $S$ ``current state'' vectors.  The cell produces two tensors as outputs: a set of $W$ output embeddings, as well as a set of $S$ ``next state'' vectors.  We denote the function going from input token embeddings to output token embeddings as the \emph{vertical} direction, and the function going from the current state vectors to the next state vectors as the \emph{horizontal} direction.  The number of state vectors $S$ and the window size $W$ are independent hyperparameters, but we set $S = W = 512$ in our experiments to simplify comparisons against baselines.
 
The \textbf{vertical direction} of the cell is an ordinary transformer layer with an additional cross-attention operation, much like a decoder layer in a standard encoder-decoder architecture \cite{vaswani2017attention}.  It does self-attention over the input tokens, and cross-attends to the recurrent states.  Unlike a typical decoder layer, we do self-attention and cross-attention in parallel.  The results of both forms of attention are concatenated together and fed into a linear projection.

The \textbf{horizontal direction} of the cell mirrors the forward direction, except that it performs self-attention over the current state vectors, and cross-attends to the input tokens.  The recurrent direction also replaces the residual connections with gates, which allows the model to ``forget'', an ability that is important for algorithmic tasks \cite{csordas2022datarouter}, or when processing long documents, where it has been central to the success of LSTMs \cite{greff2016lstm}.

Note that the presence of gates is the reason why self-attention and cross-attention are done in parallel.  Doing them sequentially, as is standard practice, would introduce a third gate in the horizontal direction, which led to worse perplexity in our experiments.

Recurrence is integrated with the sliding window attention mechanism.  Although not shown in Figure \ref{fig:recurrent-cell}, each cell also receives keys and values from the previous block as input, these are concatenated with ($\mK_e$, $\mV_e$) from the current block in order to implement sliding-window attention.

A Block-Recurrent Transformer \emph{layer} processes the blocks within a segment sequentially by stacking recurrent cells horizontally, with the ``next states'' output of the previous cell feeding into the ``current states'' input of the next cell.  In code, this is implemented as a simple for-loop over blocks.  Multiple layers can also be stacked vertically in the usual fashion.  Our experiments use a single recurrent layer, sandwiched between a number of non-recurrent layers that use sliding-window attention.
\looseness=-1

The final set of state vectors from the last block in the segment are cached, along with the keys and values, and used as the initial state for the first block on the next training step.  Every layer in the stack (both recurrent and non-recurrent) has its own cache.

\paragraph{Sharing of keys and values.} 

Keys and values are shared between the vertical and horizontal directions.  One set of keys and values ($\mK_e, \mV_e$) are computed from the input token embeddings, 
and another set of keys and values ($\mK_s, \mV_s$) are computed from the recurrent state vectors. 
Queries are not shared, so there are four separate sets of queries: $\mQ^v_e$ and $\mQ^v_s$ 
in the vertical direction, and $\mQ^h_s$ and $\mQ^h_e$ in the horizontal direction.

\subsection{State IDs and Position Bias}
\label{section:state-ids}

With a large number of state vectors, the total size of the recurrent state is far larger than that of an LSTM.  However, the same weights (projection matrices and MLP) are applied to each state vector.  Without some way to differentiate the states, the model will compute the same result for each state vector, thus negating any advantage from having multiple states.
To prevent this failure mode, we add a set of learned ``state IDs'' to the state vectors before computing the keys, values, and queries.  These ``state IDs'' allow each state vector to consistently issue different queries against the input sequence, and against other states.  State IDs are identical to learned position embeddings; we use a different name because there's no notion of ``position'' between states.

We do not add global position embeddings to the tokens, because global position embeddings don't work well for long sequences \cite{dai2019transformerXL}.  Instead, we add a T5-style relative position bias \cite{Raffel2020t5journal} to the self-attention matrix in the vertical direction.
(Although similar, T5 relative positions differ slightly from the relative positions used in the Transformer-XL paper \cite{dai2019transformerXL}.)
When the recurrent states cross-attend to input tokens, there is no position bias, because the relative distance between ``state'' and ``token'' is undefined.

We also normalize queries and keys as described in  \cite{henry2020querykeynorm}; we found that normalization improved the stability of Transformer-XL when used with a relative position bias.

\subsection{Gate Type}
\label{section:gate-type}

We experimented with two different gating mechanisms for the recurrent cell. Each state vector has its own gate, but all state vectors are updated in parallel, using the equations below.

\paragraph{Fixed gate.}  
The fixed gate uses a learned convex combination, similar to highway networks \cite{srivastava2015highway}.
\begin{align}
    \vz_t &= \mW_{z}\vh_t + \vb_{z} \\   
    \vg &= \sigma(\vb_{g}) \\
    \vc_{t+1} &= \vc_t \odot \vg + \vz_t \odot (1 - \vg)
\end{align}
where $\mW_{z}$ is a trainable weight matrix,
$\vb_{z}$ and $\vb_{g}$ are trainable bias vectors, 
$\sigma$ is the sigmoid function, 
$\vc_t$ is the cell state for the current block (i.e., the state for the block at index $t$ in the sequence of blocks),
$\odot$ is the element-wise multiplication, 
and $\vh_t$ is the current input to the gate.  In our model, $\vh_t$ is either the output of attention, in which case $\mW_{z}$ is the linear projection that feeds into the gate, or $\vh_t$ is the output of the hidden layer of the MLP, in which case $\mW_{z}$ is the final layer of the MLP.

Unlike highway networks, the bias $\vb_{g}$ is a simple learned vector of shape $\mathbb{R}^d$, which is broadcast over all state vectors, where $d$ is the state embedding dimension.  The value of $\vg$ does \textbf{\emph{not}} depend on either the current value of the state vector $\vc_t$, or on the current input $\vh_t$, and thus remains constant (i.e., fixed) after training.  The fixed gate essentially implements an exponential moving average over previous blocks.

\paragraph{LSTM gate.} The LSTM gate uses the standard combination of input and forget gates:
\begin{align}
    \label{eq:gates}
    \vz_t &= \text{tanh}(\mW_{z} \vh_t + \vb_{z}) \\
    \vi_t &= \sigmoid(\mW_{i} \vh_t + \vb_{i} - 1) \\
    \vf_t &= \sigmoid(\mW_{f} \vh_t + \vb_{f} + 1) \\
    \vc_{t+1} &= \vc_t \odot \vf_t + \vz_t \odot \vi_t
\end{align}
where $\mW_{z}, \mW_{i}, \mW_{f}$ are trainable weight matrices, and
$\vb_{z}, \vb_{i}, \vb_{f}$ are trainable bias vectors.
The LSTM gate is strictly more expressive, because the values of $\vf_t$ and $\vi_t$ depend on the current input $\vh_t$.  In our model, $\vh_t$ depends on $\vc_t$, so the LSTM gate also depends indirectly on $\vc_t$. 
LSTM gate values are thus different for each state vector, and for each block index $t$.

\subsection{Gate Initialization and Training Stability}
\label{section:gate-initialization}

We observed that training stability is quite sensitive to how the gates are initialized.  Recurrence has a failure mode where the model learns to completely ignore the recurrent state, in which case its performance reverts to that of the non-recurrent transformer.  Moreover, this situation appears to be a local optimum; once the model has reached this point, it does not recover.  We stabilize training by initializing the weights and bias to small but non-zero values, and adding a constant -1 and +1 to the input and forget gates to bias them to ``remember''.  See Appendix \ref{appendix:gate-init} for details.

\subsection{Gate Configuration}

We experimented with three different gate configurations.  

\paragraph{Dual.} 
The dual gate configuration is the one shown in Figure \ref{fig:recurrent-cell}, in which both of the residual connections in the cell are replaced with gates. The disadvantage of this configuration is that there are two gates, both of which can forget.

\paragraph{Single.} 
The single gate configuration removes the linear projection and the gate that is attached to it. Instead, the concatenation of self-attention and cross-attention is fed directly into the MLP.

\paragraph{Skip.} 
The skip configuration removes the MLP and the gate that is attached to it. This configuration is similar to the single-gate version, except that it is strictly weaker.  Instead of a two layer MLP with a very large hidden layer, it uses a linear projection with no nonlinearity.

\subsection{Placement of Recurrence and Computation Cost}

\paragraph{Single recurrent layer.} The basic version of the Block-Recurrent Transformer uses a single recurrent layer sandwiched between a number of non-recurrent transformer layers with sliding attention. We use a 12-layer model with recurrence on layer 10.  All layers have a Transformer-XL-style cache.

\paragraph{Cost of recurrence.}  During training, the 12-layer Block-Recurrent Transformer has almost exactly the same computation cost, in both parameters and FLOPS, as a 13-layer Transformer-XL model without recurrence.  The two are equivalent because the recurrent cell does almost the same operations as a conventional transformer layer, merely in the horizontal instead of the vertical direction.  

The inference cost for autoregressive decoding is also nearly identical, for the same reason.  Recurrence adds an additional attention operation per token, the cost of which is the same as self-attention in a 13th layer.


\section{Results}
\label{section:results}

We tested the Block-Recurrent  Transformer on three different data sets of long documents: PG19, arXiv, and GitHub. 
The PG19 dataset \cite{simengsun2021pg19_long_range_context} contains full-length books written prior to 1919 from project Gutenberg. 
The arXiv dataset \cite{wu2022memorizing} is a corpus of technical papers downloaded via the arXiv Bulk Data Access\footnote{\url{https://arxiv.com/help/bulk_data}}, and filtered to include only articles labeled as ``Mathematics'' and whose \LaTeX~source is available.
The GitHub dataset \cite{wu2022memorizing} is a corpus of source code from different GitHub repositories with open-source licenses.  
All of the files in each GitHub repository are concatenated together to make one long document.

The task is auto-regressive language modeling, where the goal is to predict the next token in the sequence.  We report bits-per-token numbers (i.e. $\textrm{log}_2$ perplexity; lower is better) for all models.  Further training details for each dataset can be found in Appendix \ref{appendix:training-details}. 

\begin{table}
\begin{center}
\caption{Average bits-per-token ($\textrm{log}_2$ perplexity) of each model. The recurrent models (named \namett{Rec:gate:config}) have the same computational cost as the \namett{Slide:13L} baseline, but much better perplexity. They even outperform the \namett{XL:2048} baseline, \textbf{while running more than twice as fast.}
Measured error bars on PG19 are low, between 0.002 and 0.007, but are rounded up to 0.01 to match the precision of results in the table.
Step time is for a single training step (lower is better). 
For PG19, we train both character-level (bytes) and token-level models.}
\label{table:results}

\setlength{\tabcolsep}{0.4em}
\begin{tabular}{lccccccc}
\toprule
\multirow{2}{*}{Model}  & segment  & window & step time  & \multicolumn{2}{c}{PG19} & arXiv & GitHub \\
                        & length   & length & (relative) & bytes & tokens & tokens  & tokens  \\
\midrule
\namett{XL:512}       & 512     & 512       & 0.88 & 1.01      & $3.62 \pm 0.01$      & 1.45  & 1.21 \\ 
\namett{XL:1024}      & 1024    & 1024      & 1.20 & 0.997     & $3.59 \pm 0.01$      & 1.37  & 1.08 \\ 
\namett{XL:2048}      & 2048    & 2048      & 2.11 & 0.990     & $3.58 \pm 0.01$      & 1.31  & 1.01 \\ 
\namett{Slide:12L}    & 4096    & 512       & 0.93 & 0.989     & 3.60                 & 1.43  & 1.19 \\ 
\namett{Slide:13L}    &         &           & 1.00 & 0.989     & $3.58 \pm 0.01$      & 1.42  & 1.17 \\ 
\midrule
\namett{Rec:lstm:dual}   & 4096 & 512       & 1.06 & 0.985     & $3.54 \pm 0.01$      & 1.26 & 1.01  \\ 
\namett{Rec:lstm:single} &      &           & 1.05 & 0.962     & $3.54 \pm 0.01$      & 1.29 & 1.03  \\ 
\namett{Rec:lstm:skip}   &      &           & 1.00 & 0.969     & $3.56 \pm 0.01$      & 1.31 & 1.10  \\ 
\namett{Rec:fixed:dual}   &      &          & 1.01 & 0.957     & \textbf{3.52 $\pm$ 0.01}  & 1.27 & 0.991 \\ 
\namett{Rec:fixed:single} &      &          & 1.02 & 0.966     & $3.58 \pm 0.01$      & 1.25 & 1.00  \\ 
\namett{Rec:fixed:skip}   &      &          & \textbf{0.99} & \textbf{0.952} & 3.53 $\pm$ 0.01 & \textbf{1.24}  & 
\textbf{0.976} \\ 
\midrule
\namett{Feedback:lstm:single}  & 4096 & 512 & 1.40 & 0.977          & 3.50           & \textbf{1.22}  & - \\ 
\namett{Feedback:fixed:skip}   &      &     & 1.35 & \textbf{0.935} & \textbf{3.49}  & 1.24           & - \\ 
\midrule
Memorizing Trans. 64k     & 512   & 512   & 1.94 & 0.950  & 3.53 & \textbf{1.22} & - \\
\bottomrule
\end{tabular}
\end{center}
\vskip -3ex
\end{table}

\subsection{Baselines}

We compare the Block-Recurrent Transformer to five different baselines.  The first baseline, \namett{XL:512}, establishes a reference point against which various other improvements can be compared.  It's a 12-layer Transformer-XL model with a window size of 512, and 150 million parameters.  It has 8 heads of size 128, embedding vectors of size 1024, an MLP with a hidden layer of size 4096, and the relu nonlinearity.  It uses a Transformer-XL style cache, but no sliding window, so the segment length is the same as the window size, i.e., it is trained on segments of 512 tokens. 

\namett{XL:1024} and \namett{XL:2048} are similar, but have window sizes of 1024 and 2048, respectively.  As expected, increasing the window size improves perplexity, especially on the arXiv data set.  However, these two models still have worse perplexity than the recurrent model, as well as being much slower. 

\namett{Slide:12L} is a 12-layer transformer with a window size of 512, but uses a sliding window over a segment of 4096 tokens.  This model is almost identical to \namett{XL:512}; the only difference is that the sliding window is differentiable over multiple blocks, while the Transformer-XL cache is not.  

\namett{Slide:13L} adds a 13th layer, and is directly comparable to the recurrent models in terms of both computation cost (FLOPS or step-time), number of parameters, and segment length.  Notice that adding another layer with more parameters yields a much smaller improvement than adding recurrence.

\paragraph{Relative cost.}  All five baselines, and all 6 recurrent models, have roughly the same number of parameters: between 151 million (12 layer) and 164 million (13 layer or recurrent).  The training speed (i.e. step time) of each model is shown in Table \ref{table:results} (lower is better).  Because the raw step time depends on hardware and compiler, we report numbers relative to the \namett{Slide:13L} baseline.

\paragraph{Batch Size.}
\label{sec:batch-size}

We adjust the batch size so that each model processes the same number of tokens (and thus the same amount of training data) per training step.  Thus, \namett{XL:512} (segment length 512) runs at a batch size of 256 (8 per replica), while \namett{Slide:12L} (segment length 4096) runs at a batch size of 32 (1 per replica) on PG19.

\subsection{Benefit of Recurrence}
\label{subsec:scaling}
We compare the 5 baselines to all six gate configurations for the Block-Recurrent Transformer.
The recurrent model reliably outperforms all five baselines.  The best overall configuration is \namett{Rec:fixed:skip}, which outperforms the others in 3 out of 4 cases, and comes within the margin of error in the remaining case.  This is especially notable because it is also the fastest configuration, having a slightly \emph{lower} step time and fewer parameters than \namett{Slide:13L}, because it does not have the MLP.  It is better than the 13-layer baseline by a wide margin, and it is even better than the Transformer-XL model with a window size of 2048, which runs over 2 times slower.

The other gate configurations also outperform the 13-layer baseline, but their relative ranking varies according to the dataset.  
Despite being theoretically more powerful, the \namett{LSTM} gate tends to lag behind the \namett{fixed} gate in all of our experiments.  

\paragraph{Scaling up.}

\begin{figure}[t]
  \begin{center}
  \vskip -2ex
  \includegraphics[width=0.7\textwidth]{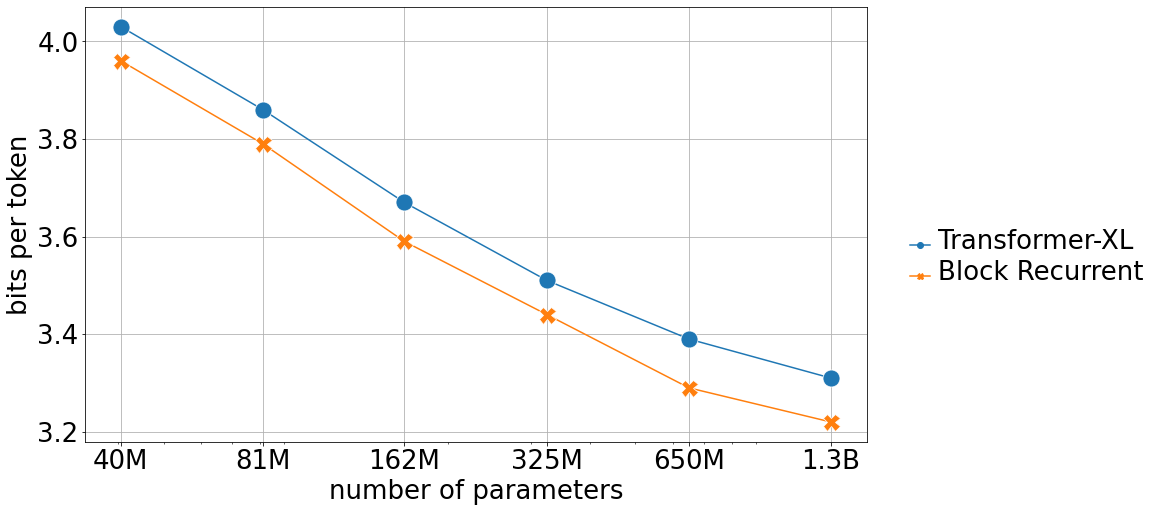}
  \caption{Scaling of the 12-layer Block-Recurrent Transformer vs 13-layer Transformer-XL on PG19. FLOPs are the same between the two models at a given parameter count. \textbf{At larger sizes, adding recurrence is equivalent to doubling the number of parameters.} Details in Appendix \ref{appendix:scaling_plot_details}.}
  \label{fig:scaling}
  \end{center}
  \vskip -3ex
\end{figure}

Figure \ref{fig:scaling} shows the effect of adding recurrence as the transformer model is scaled up and down in size.  We trained six different models on PG19, ranging in size from 40M parameters to 1.3B parameters.  For the four smaller models, we compare a 12-layer Block-Recurrent Transformer against a 13-layer Transformer-XL baseline, while for the two larger models, we compare a 24-layer Block-Recurrent Transformer, with recurrence at layers 10 and 20, against a 26-layer Transformer-XL baseline. This experiment used a cosine-decay learning rate as described in \cite{hoffmann2022training}, and a custom 32k SentencePiece vocabulary \cite{kudo18sentencepiece}.
More details are in Appendix \ref{appendix:scaling_plot_details}.

Our experiments show that recurrence provides a consistent benefit across all scales. The relative improvement actually seems to increase with the number of parameters; at larger sizes \textbf{recurrence provides a benefit which is greater than doubling the number of parameters}.  

\subsection{Ablations}

\paragraph{Multiple recurrent layers.}  Adding two recurrent layers right next to each other in the stack (layers 9 and 10) did not improve model perplexity.  Adding two layers widely separated in the stack (layers 4 and 10) did provide an improvement, but the improvement was no better than simply adding another non-recurrent layer to the stack. Previous work on Memorizing Transformers \cite{wu2022memorizing} showed a similar effect.  In our qualitative study, we saw that the model seems to use recurrence primary for long-range name lookups, much like memory. We conclude that one layer of recurrence is sufficient for the model to extract most of the benefits, although we did use two layers for our largest models.  

\paragraph{Number of recurrent state vectors.}  We trained the model with differing numbers of state vectors, from 128 to 2048.  Increasing the number of states makes a small but measurable improvement up to 1024, but the model does worse with 2048 (see Appendix \ref{appendix:number-of-states}).  We hypothesize that the model has trouble learning to use the recurrent state effectively if the state space grows too large.

\paragraph{Reducing window size.}  Reducing the size of the sliding window dramatically reduces perplexity for Transformer-XL, because it reduces the amount of context that the transformer is able to attend to.  Reducing the size of the window in a recurrent transformer has a smaller effect, because the model can use recurrence to compensate (see Appendix \ref{appendix:number-of-states}).

\subsection{Block feedback}  

Inspired by the feedback transformer \cite{fan2020feedback}, which allows all layers to attend to the topmost layer, we implemented a variation in which every layer of the transformer (not just the recurrent one) can cross-attend to the state vectors in the recurrent layer.  This variation further improves perplexity, but at a cost; step time increased by approximately 35-40\%, and the additional queries also increase the number of parameters.  Results are shown in Table \ref{table:results}, and further described in Appendix \ref{appendix:feedback}.

\subsection{Comparisons against prior published work}

\begin{table}
\begin{center}
\caption[Comparisons]{Comparison with other published work on PG19.  Fields marked - are unknown.}
\label{table:comparisons}
\setlength{\tabcolsep}{0.4em}
\begin{tabular}{lcccc}
\toprule
Model                                            & Layers         & perplexity      & parameters & vocabulary   \\
                                                 &                & word-level      &            & size     \\
\midrule
Compressive Transformer \cite{rae2020compressive}& 36             & 33.6            & -          & 32k                 \\
Routing Transformer \cite{roy2021efficient}      & 22             & 33.2            & 490M\footnotemark[\value{footnote}]       & 98k  \\
Perceiver AR \cite{hawthorne2022general}         & 60             & 28.9            & 974.6M\footnotemark[\value{footnote}]     & 32k  \\
Block-Recurrent Transformer                      & 24             & 28.46           & 650M       & 32k         \\
Block-Recurrent Transformer                      & 24             & \textbf{26.50}  & 1.3B       & 32k         \\
\bottomrule
\end{tabular}
\end{center}
\vskip -3ex
\end{table}
\footnotetext{Personal communication.}

The PG19 test set contains 6,966,499 words \cite{rae2020compressive}, which are broken into 10,229,476 tokens using a SentencePiece vocabulary, trained on PG19.  Our 24-layer 1.3B parameter model achieves 3.22 bits per token, and thus \textbf{achieves a new state of the art word-level perplexity of 26.50}  (Table \ref{table:comparisons}).  However, we note that raw perplexity numbers are not necessarily a meaningful way to compare architectures, because they depend on numerous other factors, such as the number of parameters, vocabulary, learning rate schedule, batch size, etc.; a more detailed discussion is in Appendix \ref{appendix:comparisons}. 

We were able to run a fair comparison (identical vocabulary, configuration, and hyperparameters) of the Block-Recurrent Transformer against the Memorizing Transformer \cite{wu2022memorizing}, with a memory of size 64k (Table \ref{table:results}).  The memorizing transformer is constructed similarly to our model; it has one layer which has been augmented with a mechanism that gives it the ability to attend over much longer distances.  We find that Block-Recurrence does almost as well as the Memorizing Transformer on arXiv, and does just as well on PG19, but trains almost twice as fast.  However, there are many ways of implementing approximate $k$-nearest-neighbor lookup, so relative speed will be highly implementation-dependent; our implementation runs on TPU, and does not use custom CUDA kernels.  
\looseness=-1

\subsection{Qualitative analysis}

Prior work on long-context transformers \cite{simengsun2021pg19_long_range_context, wu2022memorizing} has found that attention at long ranges is typically used to look up proper names, such as characters or places.  
We performed a qualitative analysis in an attempt to determine whether our model is using recurrence in the same way.  We selected 5 books at random from the PG19 test set, ran both the Block-Recurrent Transformer and the 13-layer Transformer-XL on each book, and then compared the cross-entropy loss for all tokens.  We sorted the results, and examined the top 4 tokens from each book with the greatest difference: the tokens for which the predictions of the recurrent model have the largest improvement over the baseline.

In 17/20 cases, the recurrent model predicted a proper name, usually with relatively high probability, that Transformer-XL was unable to predict.  In 2 cases it predicted a chapter title (having previously seen the table of contents), and in the last case, it predicted a foreign-language word that was unique to that book.  In 19/20 cases, the predicted word was nowhere within the attention window, so it must have been stored within the recurrent state (details in the appendix, Section \ref{appendix:qualitative-results}).  

In a second study, we compared the recurrent model, running normally, against a variation in which the recurrent state is cleared at the end of each 4096-token segment, instead of being cached.
Clearing the state degrades the model's ability to predict dependencies at a longer range than the segment length; typical mispredictions once again included proper names and chapter titles.  Interestingly, this study also showed that the recurrent model is able to remember the title and author of a book (which is part of the Gutenberg boilerplate at the beginning and end of each book) \textbf{across the entire length of the book -- more than 60,000 tokens.}  See Appendix \ref{appendix:clearing-state}.

A further quantitative comparison of the per-token cross-entropy between Transformer-XL and the Block-Recurrent Transformer is given in Appendix \ref{appendix:token-ce}.



\section{Discussion}

Our implementation of recurrence was inspired by the way that humans seem to process long sequences.  When a human reads a novel, they do not attempt to remember every single word in the book. Instead, a human reader will construct a mental model, or knowledge graph, which summarizes the story thus far, i.e., the names of the main characters, the relationships between them, and any major plot points. When a human reads a paragraph of text, they will parse the information in the paragraph, process and interpret the information using background knowledge from their mental model, and finally update their mental model with new information.
Our recurrent architecture loosely mimics this process. It takes a block of text, and parses it by running it through a conventional transformer stack. Tokens in the text attend to the recurrent states (i.e. the mental model), and the states, in turn, are updated by attending to the text.


Based on our qualitative analysis, it seems that the model is, in fact, using the recurrent state to summarize some of the  information about frequently occurring characters and places.  However, it does not seem to be doing much complex reasoning, as evidenced by the fact that our best performing model is the \namett{fixed:skip} configuration.  This configuration does not use a complex LSTM-style gate, which chooses to remember or forget based on its current state and inputs; instead, it simply computes an exponential moving average, not unlike some other forms of long-range approximate attention.  


Moreover, the \namett{skip} configuration cuts out the large MLP from the recurrent transformer layer.  In a vanilla transformer, removing the MLP from all layers would severely degrade the model \cite{dong2021attention:not:all:need}; those large MLPs are computing something important.  In a recurrent layer, removing the MLP makes little difference; it does not seem to be computing anything useful.
We conclude that training the recurrent layer to make full use of its capabilities for knowledge extraction and summarization will require further advances.

\subsection{Ethics} 
\label{section:ethics}

The potential negative social impacts from this work are similar to any other advance in language modelling.  Large language models could potentially be used to create disinformation and fake news, power malicious chatbots, or generate spam. The Block-Recurrent Transformer can potentially create longer documents than was previously feasible, thus expanding the range of applications in which these negative impacts could occur.  The best way to mitigate these risks is to train models that can reason about text, and flag misinformation or malicious content.

\section{Conclusion}

We have shown that when training language models on long documents, the Block-Recurrent Transformer provides a greater benefit at lower cost than scaling up the transformer model in other ways. 
Adding recurrence to a single layer has roughly the same cost as adding an additional non-recurrent layer, but results in a much larger improvement to perplexity.
We have also shown that recurrence provides a larger benefit than simply increasing the window size of attention, or increasing the number of parameters.  \textbf{Our medium-sized model has lower perplexity than a Transformer-XL model with 4 times the window size, but runs twice as fast, and our larger model outperforms a Transformer-XL model with twice the number of parameters}.

Furthermore, in contrast to some other recently proposed transformer variants, the Recurrent Transformer is very easy to implement, since it consists mostly of ordinary transformer components and RNN gates.  No custom CUDA kernels are required. Our code has been released as open source \cite{meliad2022github}.


Evaluating block-recurrent transformers on downstream tasks is an important direction for future work.  We believe that the Block-Recurrent Transformer will be most useful in situations that require long-range context; examples of potential applications include writing book reports, summarizing long news articles, code completion, or question/answering over book-length works.  There are are a number of new and emerging benchmarks that test long-range performance \cite{shaham2022scrolls, wang2022squality, tay2021long}. Previous studies have found a strong correlation between language modeling and diverse downstream tasks \cite{brown2020gpt3, srivastava2022beyond}.
\looseness=-1

Despite our initial successes, we also believe that the recurrent architecture that we present here has not yet achieved its full potential, and there are opportunities for future research and further improvements in this area.

\bibliographystyle{ieeetr}
\bibliography{recurrent_transformers}

\newpage
\begin{center}
\textbf{\huge Appendices}
\end{center}
\begin{appendices}

\section{Further Analysis.}
\label{appendix:sliding-attention}

The \emph{context length} of an autoregressive language model refers to the number of previous tokens that the model can make use of when predicting the next token. A vanilla transformer operating on segments of length $N$ has a context length of 0 for the first token, and $N-1$ for the last token, and thus has an average context length of $N/2$.  The prediction quality for a vanilla transformer is not uniform across the segment; it makes poor predictions for tokens near the beginning of the segment due to lack of context, and better predictions near the end.

With sliding window attention, each layer attends to the previous layer within a window of $W$ tokens from the current position.  The context length for a single layer is thus $W$, no matter where in the segment the token occurs.  The simple case where $W = N$ corresponds to Transformer-XL. A Transformer-XL model achieves a large improvement in average perplexity over a vanilla model, simply because it can make much better predictions for the tokens at the beginning of each segment.

In a model with multiple layers, the \emph{theoretical receptive field} (TRF) is defined as the maximum distance that information could potentially propagate through the model.  This definition is similar to ``context length'', but the TRF is ``theoretical'', because the model may have a hard time actually learning to use that much context in practice.  For example, the TRF of an LSTM is infinite, but in practice an LSTM has difficulty transmitting information over more than a few hundred tokens.  The \emph{effective context length} of an LSTM is much less than the TRF might suggest.

For a sliding-window model (or Transformer-XL), the TRF is $W \cdot L$, where $L$ is the number of layers. However, the model can still only attend directly to the previous $W$ tokens.  Although the TRF is much higher, making use of the additional context requires multiple ``hops'' of attention, which is more difficult for the model to learn.  This problem is especially acute in Transformer-XL ($N = W$), because the very first ``hop'' of attention will be to keys and values in the cache, which is not differentiable.  Our sliding window model ($N >> W$) has an identical TRF to Transformer-XL, but it can differentiate across multiple blocks, which gives it a higher effective context length in practice, and thus better perplexity.

The TRF of the block-recurrent transformer is infinite.  We also show that the effective context length also seems to be quite large in practice, since we observe cases in which the model is able to accurately predict information across distances of more than 60k tokens.

\subsection{Computational Complexity}

The computational complexity of attention in a recurrent layer is $\mathcal{O}((W^2 + S^2 + 2SW) \cdot N/W)$, where $N$ is the segment length, $W$ is the window length, and $S$ is the number of states. $N/W$ is the number of blocks, and each block does self attention $W^2$, state self-attention $S^2$, and attention between tokens/states and states/tokens $2SW$.  

The complexity of sliding-window attention is $\mathcal{O}(W^2 \cdot N/W) = \mathcal{O}(W \cdot N)$.

\subsection{Comparison between context, recurrence, and memory}

The arXiv data set contains latex code, with lots of complicated syntax that can benefit from long-range attention (e.g. theorems, jargon, citations).  As a result, adding memory, recurrence, or just increasing the window size of the Transformer-XL baseline yields a larger benefit on arXiv than it does for PG19.

There also seems to be a qualitative difference in how the models are using attention on these datasets.  We hypothesize that dealing with complicated syntax requires direct (single-``hop'') attention, which is why the \namett{XL:2048} model does well on arXiv.  In contrast, natural language novels don't have complicated syntax, but may benefit from a better understanding of subtle relationships between words (e.g. which character the word ``her'' refers to).  These subtle interactions require multi-layer, multi-``hop'' attention, which is easier to train in the more differentiable \namett{Slide:13L} model.  On PG19, \namett{Slide:13L} actually does better than \namett{XL:2048}, despite having a much shorter window size, while on arXiv, the situation is reversed. 

There may be a similar relationship between recurrence and $k$NN memory.  The Memorizing Transformer does direct, single-``hop'' attention, using $k$NN lookup.  Recurrence, in contrast, summarizes and compresses text into a set of recurrent states.  

Both the Memorizing Transformer and the Block-Recurrent Transformer achieve very similar perplexity, and based on qualitative studies, they seem to use the additional memory or states primarily for long-range name lookups.  However, recurrence may be better at capturing more subtle long-range information, like writing style, while memory is better at precision lookups of facts, like citations.  The fact that the Memorizing Transformer does better than the Block-Recurrent transformer on arXiv, but not PG19, would seem to support this hypothesis, but further experiments on downstream tasks are necessary to confirm this hypothesis.

\subsection{Comparison against Longformer}

The idea of sliding window attention was popularized by the Longformer \cite{beltagy2020longformer}, but the full Longformer is a much more complicated model than the \namett{Slide} model that we present here.

The full Longformer uses several different attention patterns, and sliding-window attention is only one of them.  Longformer also uses dilated attention and sparse global attention, both of which are implemented with custom CUDA kernels.  Moreover, LongFormer uses different window sizes in each layer, and it uses a multi-phase training regimen of pre-training and fine-tuning, following a curriculum that gradually increases window size and sequence length.  We do none of these things.

\section{Gate Initialization and Training Stability}
\label{appendix:gate-init}

We observed that training stability is quite sensitive to how the gates are initialized.  Recurrence has a failure mode where the model learns to completely ignore the recurrent state, in which case its performance reverts to that of the non-recurrent transformer.  Moreover, this situation appears to be a local optimum; once the model has reached this point, it does not recover.

Our hypothesis is that learning the recurrent transition function is a much more difficult task than learning to attend directly to the input tokens.  As a result, the vertical direction trains much faster than the horizontal direction, especially early in training.  This may lead to a situation in which the recurrent states are much less informative than the input tokens, and the model learns to ignore them.

To avoid this failure mode, gate initialization requires special care.  Moreover, proper gate initialization depends on the optimizer.  We use the Adafactor optimizer \cite{shazeer2018adafactor}, which normalizes gradients with respect to their variance, and then multiplies them by the native scale of the underlying parameter.
Thus, if a bias term is initialized to 0, its native scale will be 0, the gradient updates will be very small, and the bias will tend to remain small over the course of training.
If a bias term is initialized to 1 (which tells the forget gate to ``remember'', and is standard practice in LSTMs) then the initial updates will be large, and the model will learn to ignore the recurrent states before they have the chance to learn anything useful.

We compromise by initializing the bias terms of the gates to small but non-zero values, using a normal distribution with mean 0 and a standard deviation of 0.1.  The weight matrices of the gates are also initialized to small values, using a truncated normal distribution with a standard deviation of $\sqrt(\frac{0.1}{f_{\textrm{in}}})$ where $f_{\textrm{in}}$ is the dimension of $\vh_t$.

We add a constant of -1 and +1 to the input and forget gates (see Eq.~\ref{eq:gates}) to initially bias the gate to ``remember'' without affecting the size of the updates that Adafactor will apply. Using this initialization trick, the recurrent cell reliably learns to make use of the recurrent state.

\section{Training Details}
\label{appendix:training-details}

For PG19, we do both token-level modeling, and character-level modeling.  In our initial experiments, we use a pre-trained 32k sentencepiece vocabulary from T5 \cite{Raffel2020t5journal} for the token-level modeling.  We use the Adafactor optimizer \cite{shazeer2018adafactor}, a learning rate schedule with inverse square root decay, 1000 warmup steps, and a dropout rate of 0.05.  The learning rate is 1.0; when combined with warmup and the decay schedule, this yields an applied learning rate of 0.03, decaying to 0.0014.  This learning rate and schedule were borrowed from other language models; we did not attempt to do a hyperparameter sweep to identify the optimum learning rate and schedule.  We train for 500k steps with 32 replicas on Google V4 TPUs; training takes approximately 48 hours.  Reported results are for the "test" split.

For the later scaling experiment on PG19, we switched the learning rate schedule to cosine decay, as recommended in \cite{hoffmann2022training}, with a maximum rate of 0.01, and a minimum of 0.001.  We did a brief experiment with a learning rate of 0.02 and 0.005, before settling on 0.01.  The change in learning rate schedule resulted in a significant improvement.  We also switched from the 32k T5 vocabulary to a 32k custom sentencepiece vocabulary trained on PG19.  Our custom vocabulary has higher bits-per-token, but fewer tokens, and thus has slightly better word-level perplexity.

For arXiv, we use a pre-trained 32k vocabulary from LaMDA \cite{thoppilan2022lamda}.  Due to the large number of mathematical symbols in LaTeX, many tokens are only one character, so the bits-per-token numbers are lower than for PG19.  We dropped the learning rate to 0.5 after observing some instabilities when training on longer (4096) segment lengths.  Reported results are for the "test" split.

The GitHub dataset is much larger than PG19, and has very high variance, due to the fact that it contains code written in many different programming languages and coding styles. Consequently, there was a lot of noise in the results, which made it difficult to accurately compare models.  We reduced the noise by using a batch size that is 4x larger than for PG19. As with the PG19 scaling study, we use a cosine decay learning rate schedule, but the maximum LR is 0.005; this is half the LR used for PG19. Because of the increased batch size, these models ran for only 250k steps.  The GitHub experiment uses a pre-trained 32k vocabulary from LaMDA. Reported results are for the "validation" split. 
\looseness = -1

Due to the large number of experiments, we did not have the computational resources to run all experiments multiple times, and consequently do not provide error bars for all experiments.  However, for the headline numbers on PG19-tokens, we ran each experiment 3 times, with both different random seeds and with dataset shuffling.  Actual measured error bars on PG19 were very low, between 0.002 and 0.007.  The numbers in Table \ref{table:results} are rounded to the nearest 0.01, which means that the error bars must be rounded up to match the precision of the reported results.  E.g. for \namett{Rec:fixed:skip} on PG19-tokens, an average of 3 runs has a mean of 3.525 and a standard deviation of 0.0047; we round this \emph{up} to $3.53 \pm 0.01$.  Note that it is not possible to obtain truly accurate error bars from such a small number of runs; by rounding the error up, we provide a conservative estimate of the actual error.

\subsection{Dataset licensing and other issues.}
\label{appendix:dataset-license}

PG19 consists of works in the public domain, and consequently it is a public dataset that is freely available to other researchers.  Due to the age of the texts, some of the books do contain potentially offensive material.

The arXiv dataset consists of documents for which the author has given express permission for their work to be distributed by \texttt{arxiv.org}.  However, because the author still retains copyright, these articles cannot necessarily be redistributed in the form of a public dataset, nor will we publish a model that has been pre-trained on this data.  We obtained access to this dataset via private channels.

The Github dataset consists of code with open-source licenses, which permit the code to be downloaded, compiled, or modified.  Similar to arXiv, however, because the authors retain copyright, this code cannot necessarily be redistributed as a public dataset, nor will we publish a model that has been pre-trained on this data.  We obtained access to this dataset via private channels.

\subsection{Selection of data sets}

Having a standardized data set is import for the purpose of comparing published results, and historically, papers on long-context language modeling have used enwik8 or wiki-103 as benchmarks \cite{dai2019transformerXL}\cite{beltagy2020longformer}.  However, these datasets are not particularly good benchmarks for our purposes.

The purpose of our experiments is to see whether block recurrence can transmit information over very long distances: we show retrieval over 60k+ tokens.  We chose PG19 specifically because we believe it to be a good dataset for these sorts of experiments.  It consists only of long, book-length works, it is much larger than enwik8 or wiki-103, it is publicly available, and has been cited in other published work.  Arxiv and github are (sadly) not public, but they similarly have long documents in the 50k+ token range.

Enwik8 is not a corpus of long articles.  In fact, it doesn't even split the text into separate articles at all; it's just a single text dump that concatenates a bunch of short unrelated articles together.  Attempting to split it on article boundaries yields a data set in which the majority of ``articles'' are merely stubs, with HTML boilerplate and no actual text.  Enwik8 is a fine benchmark for data compression, which was the purpose for which it was originally intended, but it is less than ideal for long-range language modeling. 

Wiki-103 is significantly better, because it does break the text into articles, and it eliminates the boilerplate, but the average length is still only 3.6k tokens per article, which is less than the segment length used in our experiments, and much shorter than the 50k-100k tokens of PG19.

\subsection{Comparisons with previously published results.}
\label{appendix:comparisons}

It is well-known that transformer perplexity scales with the number of parameters. However, the choice of vocabulary, learning rate schedule, batch size, number of training steps, and optimizer also make a large difference to the final headline perplexity numbers.  The Chinchilla scaling study \cite{hoffmann2022training} demonstrates that a change to the learning rate schedule can have a large effect; we also observed improvements from a change to vocabulary as well.  Not all vocabularies are created equal, even for vocabularies which have the same size, and are trained primarily on English-language text.  Published perplexity numbers between different models cannot be meaningfully compared unless all other variables are strictly controlled.

\section{Window Size and Number of Recurrent States}
\label{appendix:number-of-states}

Ablation results for the window size is is shown in Table \ref{table:statewin} (a).  Decreasing window size leads to worse perplexity in both the recurrent model and Transformer-XL, but the penalty is smaller for the recurrent model.

Ablation results for the number of recurrent states is shown in Table \ref{table:statewin} (b). Increasing the number of recurrent states makes a small but measurable improvement up to 1024 states, but is worse at 2048 states.  The window size was 512 for this experiment.

\begin{table}[h]
\caption{Changing the window size (a) and number of recurrent states (b) on PG19.}
\label{table:statewin}

\begin{tabular}{ccc}
\setlength{\tabcolsep}{0.4em}
\begin{tabular}{lcc}
\toprule
Window size & \namett{Rec:fixed:skip} & \namett{Slide:12L}   \\
\midrule
128          &  3.58                   &  3.69               \\
256          &  3.54                   &  3.64               \\
512          &  3.53                   &  3.60               \\
\bottomrule
 & \\
 & \\
\end{tabular}
& &
\begin{tabular}{lc}
\toprule
Number of states  & \namett{Rec:fixed:skip} \\
\midrule
128          &  3.54                   \\
256          &  3.535                  \\
512          &  3.53                   \\
1024         &  3.51                   \\
2048         &  3.55                   \\
\bottomrule
\end{tabular}
\end{tabular} 
\vskip -2ex
\end{table}

\section{Block-Feedback}  
\label{appendix:feedback}

In the block-feedback variation of our model, the entire stack of 12 layers is applied to the first block of tokens.  The recurrent state for that block is then extracted from the recurrent layer, and the state is broadcast to all other layers when processing the next block. Because the recurrent layer is placed high in the stack, this means that the lower layers of the transformer can cross-attend to a higher layer, which is computationally more powerful, much like the feedback transformer \cite{fan2020feedback}.  Results are shown in Table \ref{table:results}.

Recurrence with feedback is significantly more expensive than the non-feedback version, because all 12 layers now have a cross-attention module, instead of just the recurrent layer.  
In our experiments, feedback increased the step time by approximately 35-40\%, and the additional queries also increase the number of parameters.

Adding feedback improves perplexity in most cases, but the improvement seems to depend on the data set.  The effect of feedback also depends on the gate configuration.  In particular, block feedback dramatically improves the performance of the LSTM gate.  This could be because the recurrent states, and thus the gate, get a gradient from all all layers of the transformer, instead of just one.


\section{Scaling Plot Details}
\label{appendix:scaling_plot_details}

\begin{table}[ht]
\begin{center}
\caption{Bits per token on PG19 at various model scales. The data in this table was used for the scaling plot in Figure \ref{fig:scaling}.}
\label{table:appendix_scale_table}
\setlength{\tabcolsep}{0.4em}
\begin{tabular}{rrrrrrr}
\toprule
    & 40.6M   &  81.2M & 162M & 325M & 650M & 1.3B \\
\midrule
\namett{Rec:fixed:skip}  & 3.96 & 3.79 & 3.59 & 3.44 & 3.29 & 3.22 \\
\namett{XL:512:13-layer} & 4.03 & 3.86 & 3.67 & 3.51 & 3.39 & 3.31 \\
\bottomrule
\end{tabular}
\end{center}
\end{table}

Performance on large text datasets, such as PG-19, is highly correlated with the number of trainable parameters; larger models tend to perform better. However, training large models can be expensive, and not all researchers have access to the necessary amount of compute to beat our new state of the art, or even to reproduce our results.  

Table \ref{table:appendix_scale_table} provides the numeric results from our scaling study, which covers scales from small 40M parameter models that can be easily trained on a single machine, to our largest 1.3B parameter model.  Configuration files for all scales are provided in the open source release.  Our scaling strategy is to increase the dimensions of the various parts of the transformer: embedding size, MLP hidden layer size, number of heads, head size, and number of layers.  Each factor of 2 increase in the number of parameters scales either one or two of these dimensions.

\section{Qualitative Analysis Results}
\label{appendix:qualitative-results}

\begin{figure}[h]
  \begin{center}
  \vskip -2ex
  \includegraphics[width=5.5in]{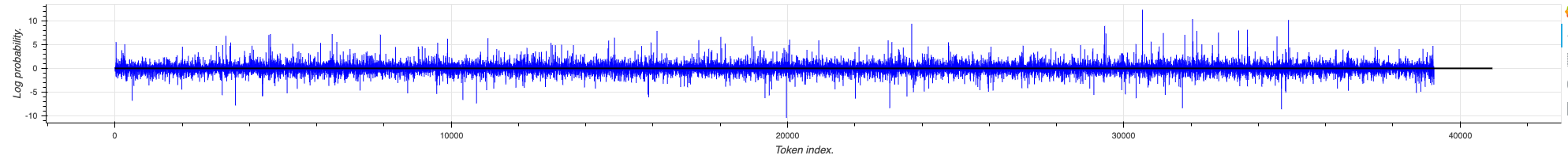}
  \vskip -2ex
  \caption{Difference in per-token cross-entropy loss.}
  \label{fig:token-difference}
  \end{center}
  \vskip -1ex
\end{figure}

The following are excerpts from our qualitative study.  We selected five books at random from PG19 test set, and ran two different models on each book.  The first model is a 13-layer Transformer-XL, and the second is the \namett{Rec:fixed:skip} configuration of the Block-Recurrent Transformer.  For each token, we compute the difference between the cross-entropy loss (i.e., the negative log likelihood (NLL)) output by both models, and and then sort the results.  

Figure \ref{fig:token-difference} shows an example of the per-token difference in NLL between the two models on the first book; the x-axis is the index of the token.  On average, the recurrent model does slightly better than Transformer-XL, but it does not necessarily make a better prediction for any individual token.

The following excerpts show the top 4 tokens where the Block-Recurrent Transformer made a better prediction than Transformer-XL; these tokens correspond to spikes in Figure \ref{fig:token-difference}.  We show the token number, the NLL returned by the recurrent model, the NLL returned by Transformer-XL, and an excerpt of text, with the token itself marked with \texttt{|token|}.  Almost all of the top tokens are proper names of characters and places. In all cases except one, the mispredicted name does not appear within the attention window of the previous 512 tokens.  These names are thus invisible to Transformer-XL, but visible to the recurrent model. 

Note that these are not cherry picked examples; the five books are chosen at random.  Moreover, the same pattern still holds if the search is expanded to the top 40 tokens for each book.  In fact, even the names are often the same; Transformer-XL often seems to mispredict the same names over and over again; these are likely the names of main characters.  

Sorting the other way, to show the top tokens where Transformer-XL does better than the recurrent model, does not show the same pattern.  There are still plenty of proper names, but it is usually cases where both models fail to predict the name.  Moreover, the names are mixed with more common words as well.

\vskip 2ex
\textbf{\emph{Memoirs of Marie Antoinette Queen Of France Vol. 7}.}

\begin{verbatim}
(30555, 0.3696011, 12.688916)
physician's house to make inquiries as to  the cause of so long an absence.
G|omin| and Larne had not yet ventured to  follow this advice, when next
\end{verbatim}
  
\begin{verbatim}
(32043, 0.20177796, 10.513703)
with the autopsy arrived at the  outer gate of the Temple.  These were
Dum|ang|in, head physician of the  Hospice de l'Unite; Pelletan,
\end{verbatim}
  
\begin{verbatim}
(34896, 1.3752508, 11.534926)
who had acted as courier to Louis  XVI. during the flight to Varennes, 
and Tur|gi|, who had waited on the  Princesses in the Temple.  It was
\end{verbatim}
  
\begin{verbatim}
(34896, 1.3752508, 11.534926)
.    In all the evidence there appeared but two serious facts, attested 
by  Latour-|du|-Pin and Valaze, who deposed to them because they could
\end{verbatim}

Notes: ``Gomin'' appears multiple times in the top 40 results.

\vskip 2ex
\textbf{\emph{An Unsentimental Journey through Cornwall}}

\begin{verbatim}
(983, 0.80441666, 11.626528)
OUGH CORNWALL          [Illustration: FALMOUTH, FROM FLUSHING.]
|DAY| THE FIRST      I believe in holidays. Not in a frantic rushing 
\end{verbatim}
  
\begin{verbatim}
(23606, 0.655162, 11.265186)
there was not the slightest use  in getting up, I turned round and took
another sleep.          |DAY| THE FIFTH      "Hope for the best, and be 
\end{verbatim}
  
\begin{verbatim}
(11297, 0.405902, 10.426135) 
ball.    "Ma'am, if you go slow and steady, with me before and Cur|gen|ven 
behind,  you'll _not_ fall."    Nor did I. I record it
\end{verbatim}
  
\begin{verbatim}
(38021, 1.3457009, 11.167879) 
our journey; going over the same ground which  we had traversed already, 
and finding Praden|ack| Down as bleak and  beautiful as ever. Our first 
\end{verbatim}

Notes: The word ``DAY'' appears earlier in the table of contents, but not within the previous 512 tokens.  ``Curgenven'' appears multiple times in the top 40.

\vskip 2ex
\textbf{\emph{The Good Hope} by Herman Heijermans Jr.}

\begin{verbatim}
(3975, 0.7425603, 15.8969145)
to us."    It matters nothing that this gospel of Life has often been  
preached. He|ij|ermans has caught the spirit of it as well as the  letter. 
\end{verbatim}

\begin{verbatim}
(7385, 0.26241615, 15.000762)
must scratch the stones.    CLEM. Tomorrow afternoon, then.    COB. T|ja|! 
I'll be here, then. Good day, Miss. [To Barend.] Good
\end{verbatim}
 
\begin{verbatim}
(42638, 0.73628587, 15.407666)
hundred guilders. Bejour. [Rings off;  at the last words Kne|irt|je has 
entered.]    KNEIRTJE. [Absently.] I----[
\end{verbatim}
  
\begin{verbatim}
(25798, 0.12310324, 14.402218)
They exeunt, dragging Barend.]    KNEIR. Oh, oh----    TRU|US|. [With 
anxious curiosity, at side door.] What was the matter,  Kneir?
\end{verbatim}
 
Notes: The word ``Tja'' is not a proper name, but is a Dutch word that is likely unique to this particular book.  It appears multiple times in the top 40 results. The name ``Truus'' appears multiple times in the top 40.

\vskip 2ex
\textbf{\emph{Travels in Morocco Volume 2} by James Richardson}

Notes: The second example in this list is not a good one, because both models mispredict the name.  We thus include a fifth example as well.  The word ``Toser'' appears frequently in the top 40, and is also the only case where the proper name appears within the attention window of 512 tokens.

\begin{verbatim}
(61049, 2.0328524, 18.696453)
, some of them as black as [Redacted]s. Many people in  T|oser| have sore 
eyes, and several with the loss of one eye, or nearly  so; opthal
\end{verbatim}
  
\begin{verbatim}
(63453, 9.254061, 24.413633)
Tunis;" but the restrictive system established by the Turks  during late 
years at Ghad|umes|, has greatly damaged the trade between the  Jereed and 
\end{verbatim}  
  
\begin{verbatim}
(66149, 0.8768588, 14.105822)
enterprizing fellow,  worthy of imitation. He calculated the distance from 
Ghabs to T|oser| at  200 miles. There are a number of towns in the 
\end{verbatim}  
  
\begin{verbatim}
(64161, 1.0309317, 14.152167)
and ourselves went to Wedyen, a town and  date-wood about eight miles 
from T|oser|, to the left. The date-grove is  extensive, and there are
\end{verbatim}  
  
\begin{verbatim}
(27282, 0.04982663, 12.551973)
, is a very ancient city, situate upon the right bank of the  river 
Boura|gra|g, and near its mouth. This place was captured in 1263, by
\end{verbatim}

\vskip 2ex
\textbf{\emph{India's Love Lyrics by Laurence Hope}}

\begin{verbatim}
(15694, 0.07477246, 11.22447)
rieving            A wasted chance.  Fate knows no tears.       Verses: 
Fa|iz|Ulla       Just in the hush before dawn     A little wistful wind is 
\end{verbatim}  
  
\begin{verbatim}
(3135, 0.0126608405, 9.759871)
afloat     In the little noontide of love's delights            Between 
two Nights.            Val|go|vind's Boat Song       Waters glisten and 
sunbeams quiver,
\end{verbatim}  
  
\begin{verbatim}
(23183, 0.1419289, 9.481476)
sleep, the touch of your lips on my mouth.            His Rubies: Told by 
Val|go|vind       Along the hot and endless road,       Calm and erect, 
with haggard eyes,'
\end{verbatim}  
  
\begin{verbatim}
(26886, 0.13428347, 9.46193)
off sheath,       And find there is no Rival like the Past.     Verse 
by T|aj|Mahomed       When first I loved, I gave my very soul    Utterly 
\end{verbatim}

\subsection{Clearing the recurrent state}
\label{appendix:clearing-state}

\begin{figure}[ht]
  \begin{center}
  \vskip -2ex
  \includegraphics[width=5.5in]{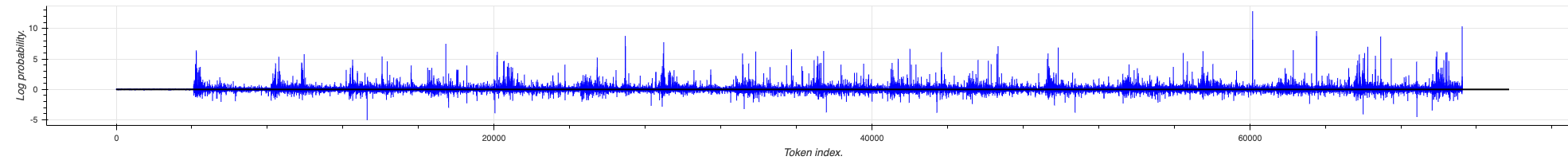}
  \vskip -2ex
  \caption{Difference in per-token cross-entropy loss with state clearing.}
  \label{fig:baby-mine}
  \end{center}
  \vskip -1ex
\end{figure}

Our second qualitative study is structured similarly to the first, except that instead of comparing two different models, we compare two different runs of the same model: the \namett{Rec:fixed:skip} configuration.  The first run processes the book normally, while the second run clears the recurrent states at the beginning of each 4096-token segment.  In the second run, the model can use recurrence within a segment to look beyond the local attention window of 512 tokens, but it cannot use recurrence to carry information from one segment to the next. 

This experiment is somewhat cleaner than the first, because both runs are done with the same pre-trained model, which has the same parameters. Figure \ref{fig:baby-mine} shows the difference in per-token cross-entropy loss for the first book.  There is no difference between the two runs for the first segment, and the biggest differences in subsequent segments are often clustered near the start of each new segment.

The overall pattern is very similar to the first qualitative experiment: most of the tokens involve proper names.  We verified that in most cases, the mispredicted name not only does not occur within the 512-token attention window, but does not occur within the 4096-token segment.  In addition to proper names, chapter titles and illustration captions occur frequently within the top 40 results; the recurrent model seems to be remembering these  from a previous occurrence in the table of contents. 

\textbf{Perhaps most interestingly, in two of the books, one of the highest ranked mispredictions was the title and author of the book itself.}  The Gutenberg project inserts boilerplate at both the beginning and end of each book; the title and author are listed multiple times at the beginning, and once at the end.  This experiment thus shows that the model is able to ``remember'' this information in the recurrent state, across a distance of 60,000 tokens or more. 

\vskip 2ex
\textbf{\emph{Baby Mine}, by Margaret Mayo}

Note that the 2nd misprediction \textbf{is the title of the book itself}, at token number 71,244.  

\begin{verbatim}
(60153, 0.00438364, 12.798895)
nerves, but you  needn't worry, I've got everything fixed. 
Donneg|hey| sent a special  officer over with me. He's outside 
\end{verbatim}

\begin{verbatim}
(71244, 2.3727102, 12.660636)
sunlight and shadows of his  ample, well kept grounds.      End of the 
Project Gutenberg EBook of| Baby| Mine, by Margaret Mayo    ***
\end{verbatim}

\begin{verbatim}
(63536, 4.132567, 13.688847)
Alfred, with the air of a  connoisseur.    "She sure is," admitted
Don|neg|hey, more and more disgruntled as he felt  his reputation for
\end{verbatim}

\begin{verbatim}
(26947, 3.7521973, 12.516711)
with a sigh of thanksgiving he hurried upstairs to his unanswered mail.        
CHAPTER XIII    When Alfred| Hardy| found himself on the train bound for 
\end{verbatim}

\vskip 2ex
\textbf{\emph{The 'Patriotes' of '37} by Alfred D. Decelles}

\begin{verbatim} 
(38759, 0.85374373, 11.140007)
24, 125,  126.    Cote, Dr Cyri|le|, 89, 108, 118, 120;
\end{verbatim}

\begin{verbatim}
(40181, 0.35475963, 10.291676)
17-26, 129-30.    Nelson, Dr Wolf|red|, a follower of Papineau, 37, 60
\end{verbatim}

\begin{verbatim}
(28756, 0.0010796335, 9.256147)
on a well-reasoned plan of  action.  Most of the leaders--Wolf|red| Nelson, 
Thomas Storrow Brown,  Robert Bouchette, and Amury Girod--were
\end{verbatim}

\begin{verbatim}
(28772, 1.7782648, 10.611834)
leaders--Wolfred Nelson, Thomas Storrow Brown,  Robert Bouchette, and
Am|ury| Girod--were strangers to the men under  their command; and none of
\end{verbatim}

\vskip 2ex
\textbf{\emph{Another Brownie Book}, by Palmer Cox}

In this case, \textbf{the top 4 results include the author of the book}.  This book has a lot of illustrations which are listed in the table of contents, and make up many of the other results in the top 40.

\begin{verbatim}
(16442, 0.087840006, 8.706101)
[Illustration]    [Illustration]          [Illustration]    THE BROWNIES 
AND THE TUG|BO|AT.      [Illustration]    While Brownies strayed along a 
\end{verbatim}

\begin{verbatim}
(24225, 0.04784866, 8.266858)
[Illustration]            End of the Project Gutenberg EBook of Another 
Brownie Book, by Palmer| Cox|   *** 
\end{verbatim} 

\begin{verbatim}
(24224, 5.651471, 11.840027)
Illustration]    [Illustration]            End of the Project Gutenberg 
EBook of Another Brownie Book, by| Palmer| Cox    *** 
\end{verbatim}

\begin{verbatim}
(4460, 1.0682098, 6.6397715)
To secret haunts without delay.    [Illustration]          [Illustration]
THE BROWNIES AT| AR|CHERY.      [Illustration]    [Illustration]
\end{verbatim}
 
\vskip 2ex
\textbf{\emph{The Life Of Thomas Paine Vol. II.} (of II) by Moncure Daniel}

By the end of the book, the recurrent model is quite sure that "Paine" is a main character, but not if its memory keeps getting erased. 

\begin{verbatim}
(170457, 0.98941976, 12.209277)
they could. The scandal branched into  variants. Twenty-five years later 
pious Grant| Thor|burn promulgated that  Paine had run off from Paris
\end{verbatim}

\begin{verbatim}
(120592, 0.6974209, 10.396527)
my friends and accept the same to yourself."    As the Commissioners did 
not leave when they expected,| Paine| added  several other letters to
\end{verbatim}

\begin{verbatim}
(169152, 0.18432029, 9.516743)
whose composition the farrier no doubt supposed  he had paid the editor 
with stories borrowed from "Old|ys|," or not  actionable. Cheetham 
\end{verbatim} 

\begin{verbatim}
(139870, 0.53406477, 9.646917)
nor was any letter received from  him. This was probably the most important
allusion in a letter of| Paine|,  dated New York, March 1, 1804, to "C
\end{verbatim}

\begin{figure}[ht]
\begin{center}
\includegraphics[width=\textwidth]{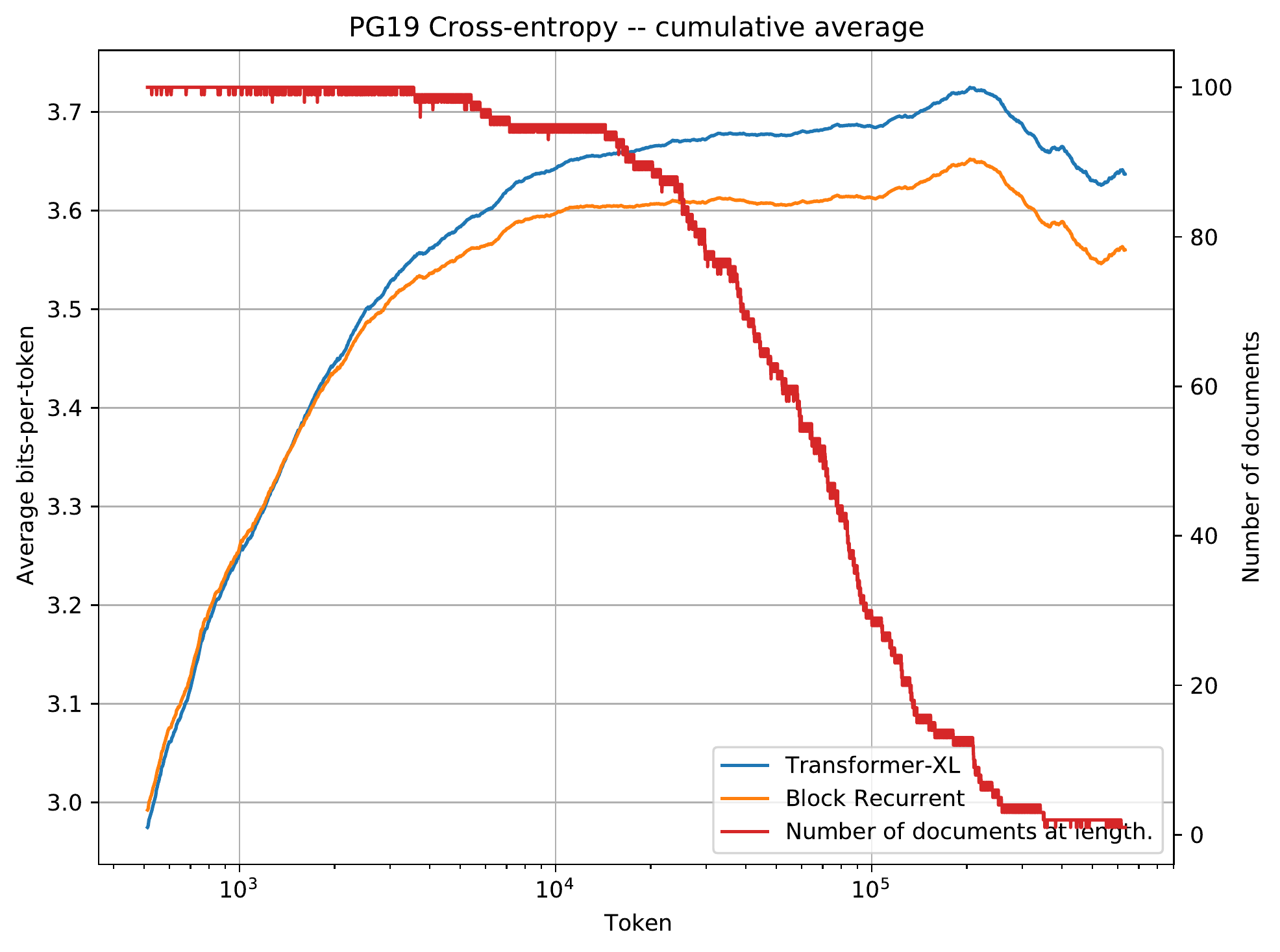}
\caption{Cumulative cross-entropy on PG19 of a 13-layer Transforemer-XL and Block Recurrent model. Though comparable at the first few thousand tokens, the recurrent model performs better at longer sequences. In red we show the number of documents at a given token length.
}
\label{fig:pg19-token-ce}
\end{center}
\end{figure}

\section{Token level cross-entropy}
\label{appendix:token-ce}

In addition to qualitative studies comparing a single document, we also compared the average bits-per-token over all documents in the PG19 test set.  It has been observed that in vanilla transformer architectures, this token-wise cross-entropy often diverges at long segment lengths \cite{press2022train}.

In Figure~\ref{fig:pg19-token-ce} we plot the cumulative cross-entropy, which is the average bits-per-token (i.e. log$_2$ perplexity) averaged up to the given length, and we compare the Block Recurrent Transformer against the Transformer-XL baseline. Performance of the two architectures is comparable for the first few thousand tokens, but the recurrent architecture clearly outperforms Transformer-XL at longer document lengths. 

\section{Vocabulary Ablation Experiments}



The authors of the PG-19 dataset propose \textit{word-level perplexity} (WLP) as a measure of model performance \cite{rae2020compressive}.  
However, it is only possible to compute WLP if there is an accurate count of the number of words.  A published word count is provided for the PG19 data set \cite{rae2020compressive}, but a word count is not available for the other data sets in this paper, such as arXiv and github.  There is not even a clear definition of what constitutes a ``word'' in source code or \LaTeX\  documents, which have a lot of symbols.  Consequently, we do not provide WLP for these data sets.

WLP also glosses over some details of tokenization; in particular, the word count ignores whitespace.  SentencePiece \cite{kudo18sentencepiece} is often advertised as being a subword tokenizer that preserves whitespace, but in fact that is only partially true.  Depending on how it is configured, SentencePiece may normalize text by mapping multiple whitespace characters (e.g. tab) to space, merge internal whitespace characters,  or strip all newlines from the text.

Thus, depending on vocabulary, a model trained with a SentencePiece tokenizer may or may not be predicting whitespace and newlines.  If it does predict them, then that whitespace is not captured by the word count, which can affect the WLP numbers.

Measuring bits-per-character (BPC) rather than WLP is a potentially simpler and fairer comparison that works with any data set and tokenizer, but it runs into the same problem with whitespace.  If the tokenizer merges or strips whitespace from the input text, as many SentencePiece vocabularies do, then it is not possible to get a consistent character count across different vocabularies. 

\subsection{Effect of vocabulary on model perplexity}

The vocabulary itself can also make a significant difference.  We argue that the tokenizer and vocabulary should be considered as part of the model, and comparisons using either WLP or BPC are only valid between models that use the same tokenizer and vocabulary.  Using WLP to judge model innovations without taking into account other differences can lead to incorrect conclusions.  We give an empirical example of how using different vocabularies can lead to significant differences in WLP.

The total cross-entropy loss $L_S = -\sum_{t=0}^{N_S} \textrm{log}(p_t|p_{<t})$ where $S$ is the chosen tokenization scheme and $N_S$ is the number of tokens in the data when tokenized by $S$.
Models using the same tokenization scheme can be compared by their \textit{token-level perplexity} $\text{PPL}_S = \exp(\frac{L_S}{N_S})$ where $\frac{L_S}{N_S}$ is the \textit{average loss}. Lower perplexity is better.

WLP is defined to be $\exp(\frac{L_S}{N_W})$ where $N_W$ is the number of "words" as defined by the authors of the PG-19 datasets. 
For the PG-19 test set, $N_W =  6,966,499$ \cite{rae2020compressive}. 
Thus, WLP is simply a rescaling of the average loss, i.e. WLP =  ${\exp(\frac{L_S}{N_S} \frac{N_S}{N_W})}$, where we'll refer to $\frac{N_S}{N_W}$ as the \textit{scale factor} -- the number of tokens per word. 

\subsubsection{Vocabulary size}

The size of the vocabulary has three obvious effects on the model.  First, a larger vocabulary has more parameters, and thus can store more information in the embedding table.  

Second, a larger vocabulary will typically have longer and rarer words, so the average length of each token (characters-per-token) will also be longer, and the \emph{scale factor} (tokens-per-word) will be smaller.  The context length of the model \emph{in tokens} is usually fixed, so having longer tokens means that the model can attend over longer distances in the input text.  

Third, having longer tokens also means that the model will process more of the training data on each training step (again because the number of tokens per step is fixed), and thus complete more epochs over the data set. 

\subsubsection{Vocabulary quality}

Even if two vocabularies are the same size, there may be a difference in vocabulary quality.  For example, if a multi-lingual vocabulary is used on English-language text, then only a portion of its total capacity will be used to capture English words, and it will behave much like a smaller vocabulary.

A more subtle issue is that there are many ways in which text can be tokenized.  A tokenizer which does a good job of capturing natural semantic segmentation into subwords can be expected to perform better than one that doesn't, e.g. ``\texttt{walk|ed}'' vs. ``\texttt{wal|ked}.''

\subsection{Experimental results}

We train the \namett{Rec:fixed:skip} model with a segment length of 4096 for 500k steps employing an inverse square root decay schedule while only varying the choice of vocabulary (all other hyperparameters are as in Section \ref{section:results}).

The byte-level vocabulary does not use a tokenizer, and operates on the raw ASCII text.  The T5 vocabulary~\cite{Raffel2020t5journal} was trained on multi-lingual data from Common Crawl (commoncrawl.org), consists of 32k tokens, and does not preserve whitespace.  We also test on a 32k sized version of the LaMDA vocabulary~\citep{adiwardana2020towards}.  To test the effect of vocabulary size, we also train a number of SentencePiece models on the PG19 training set with sizes ranging from 512 to 128k. 

\begin{figure}[ht]
\begin{center}
\includegraphics[width=0.9\textwidth]{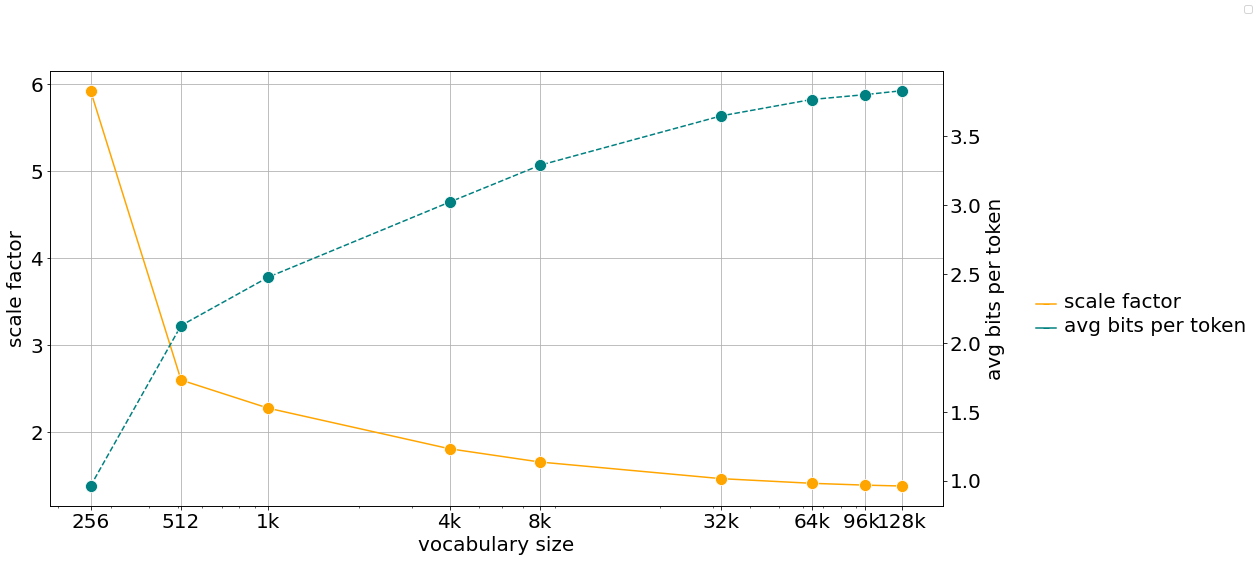}
\caption{Word-perplexity scale factor and average bits per token for sentencepiece vocabularies trained on PG19 with different sizes.}
\label{fig:vocab-comparison}
\end{center}
\end{figure}

Results are shown in Table \ref{table:vocab_comparisons} and Figure \ref{fig:vocab-comparison}.
For SentencePiece vocabularies trained on PG19, we observe that a larger vocabulary has a significant effect on the scale factor, although the effect starts to diminish with more than 32k entries (see Figure \ref{fig:vocab-comparison}).  The average bits per token increases as the tokens become longer. 

\begin{table}
\begin{center}
\caption{Test perplexity of the \namett{Rec:bias:skip} model when using different tokenizers.  Results are given in bits-per-token, so WLP is $2^{(\textrm{bits-per-token}\ \cdot\ \textrm{scale factor})}$.  STE stands for Tensorflow's deprecated SubwordTextEncoder as used by other baselines; SPM stands for SentencePiece model \cite{kudo18sentencepiece}.}
\label{table:vocab_comparisons}
\setlength{\tabcolsep}{0.4em}
\begin{tabular}{cccccc}
\toprule
vocab   & size   & token count & scale factor & avg bits-per-token & WPL \\
\midrule
T5 \citep{Raffel2020t5journal}
         & 32k  &  10,523,460 & 1.5106 & 3.525 & 40.08 \\
LaMDA \citep{thoppilan2022lamda}
         & 32k  &  12,053,681 & 1.7302 & 3.121 & 42.23 \\
PG19 STE & 32k  &  11,097,364 & 1.5930 & 3.352 & 40.49 \\
PG19 SPM & 32k  &  10,229,476 & 1.4684 & 3.647 & 40.93 \\
\midrule
PG19 SPM & 128k &  9,638,472 & 1.3835 & 3.830 & 39.37 \\
PG19 SPM & 96k  &  9,711,259 & 1.3940 & 3.801 & \textbf{39.36} \\
PG19 SPM & 64k  &  9,856,687 & 1.4149 & 3.766 & 40.18 \\
PG19 SPM & 32k  &  10,229,476 & 1.4684 & 3.647 & 40.93 \\
PG19 SPM &  8k  &  11,559,455 & 1.6593 & 3.289 & 43.94 \\
PG19 SPM &  4k  &  12,622,101 & 1.8118 & 3.021 & 44.43 \\
PG19 SPM &  1k  &  15,865,499 & 2.2774 & 2.477 & 49.91 \\
PG19 SPM & 512  &  18,124,332 & 2.6016 & 2.125 & 46.16 \\
bytes    & 256  &  41,288,269 & 5.9267 & 0.960 & 51.61 \\
\bottomrule
\end{tabular}
\end{center}
\end{table}

\end{appendices}

\end{document}